\documentclass{article}
\usepackage[final, dblblindworkshop]{neurips_2026}
\workshoptitle{Africa in AI at NeurIPS 2026}
\usepackage[utf8]{inputenc}
\usepackage[T1]{fontenc}
\usepackage{hyperref}
\usepackage{url}
\usepackage{booktabs}
\usepackage{amsfonts}
\usepackage{nicefrac}
\usepackage{microtype}
\usepackage{xcolor}
\usepackage{tikz}
\usepackage{pifont}
\usepackage{parskip}
\usetikzlibrary{arrows.meta,backgrounds,positioning,fit,calc}
\usepackage{natbib}

\usepackage[most]{tcolorbox}
\usepackage{listings}
\usepackage[table]{xcolor}
\usepackage{fancyhdr}
\definecolor{Cgray}  {HTML}{D3D1C7}
\definecolor{Cgdk}   {HTML}{444441}
\definecolor{Camber} {HTML}{FAC775}
\definecolor{Cadk}   {HTML}{633806}
\definecolor{Cteal}  {HTML}{9FE1CB}
\definecolor{Ctdk}   {HTML}{085041}
\definecolor{Cblue}  {HTML}{B5D4F4}
\definecolor{Cbdk}   {HTML}{0C447C}
\definecolor{Cgreen} {HTML}{C0DD97}
\definecolor{Cgdk2}  {HTML}{27500A}
\definecolor{Cpurp}  {HTML}{CECBF6}
\definecolor{Cpdk}   {HTML}{3C3489}
\definecolor{Ccoral} {HTML}{F5C4B3}
\definecolor{Ccdk}   {HTML}{712B13}

\lstdefinestyle{promptstyle}{
    basicstyle=\ttfamily\small,
    breaklines=true,
    breakatwhitespace=true,
    columns=flexible,
    keepspaces=true,
    showstringspaces=false,
    upquote=true,
    literate={`}{{\textasciigrave}}1
}

\newtcblisting{promptbox}[2][]{
    listing only,
    listing style=promptstyle,
    title=#2,
    colback=gray!5,
    colframe=black!60,
    colbacktitle=black!75,
    coltitle=white,
    fonttitle=\bfseries,
    boxrule=0.5pt,
    arc=2pt,
    left=6pt,
    right=6pt,
    top=4pt,
    bottom=4pt,
    enhanced,
    #1
}

\tikzset{
  arr/.style={-{Stealth[length=5pt,width=4pt]},thick,color=Cgdk},
  GB/.style={draw=Cgdk, fill=Cgray,  rounded corners=4pt, text=Cgdk,
             align=center, inner sep=5pt, text width=#1 cm, font=\footnotesize},
  AB/.style={draw=Cadk, fill=Camber, rounded corners=4pt, text=Cadk,
             align=center, inner sep=5pt, text width=#1 cm, font=\footnotesize},
  TB/.style={draw=Ctdk, fill=Cteal,  rounded corners=4pt, text=Ctdk,
             align=center, inner sep=5pt, text width=#1 cm, font=\footnotesize},
  BB/.style={draw=Cbdk, fill=Cblue,  rounded corners=4pt, text=Cbdk,
             align=center, inner sep=5pt, text width=#1 cm, font=\footnotesize},
  RB/.style={draw=Cgdk2,fill=Cgreen, rounded corners=4pt, text=Cgdk2,
             align=center, inner sep=5pt, text width=#1 cm, font=\footnotesize},
  PB/.style={draw=Cpdk, fill=Cpurp,  rounded corners=4pt, text=Cpdk,
             align=center, inner sep=5pt, text width=#1 cm, font=\footnotesize},
  CB/.style={draw=Ccdk, fill=Ccoral, rounded corners=4pt, text=Ccdk,
             align=center, inner sep=5pt, text width=#1 cm, font=\footnotesize},
  CT/.style={draw=Cgdk, fill=#1!8, dashed, thick,
             rounded corners=6pt, inner sep=8pt},
  SL/.style={font=\bfseries\footnotesize, text=Cgdk, anchor=west},
  HL/.style={font=\bfseries\scriptsize},
}

\title{Show Your Work in Swahili: Reinforcement Learning for
Native-Language Mathematical Reasoning}

\author{%
    Jessica Lundin\thanks{Equal contribution. Contact: jessica.lundin@gatesfoundation.org} \\
    Gates Foundation \\
    \And 
    Katherine Rosenfeld\footnotemark[1] \\
    Gates Foundation \\
    \And
    David Adelani \\
    McGill University
    \And 
    Geoffrey Mosoti Nyakiongora \\
    Gates Foundation
    \And 
    Halfan Ngowo \\
    Gates Foundation
    \And
    Sloan Coffin \\
    Gates Foundation
    \And 
    Ralph Abboud \\
    RenPhil
    \And 
    Guillaume Chabot-Couture \\
    Gates Foundation
}

\begin{document}

\maketitle

\begin{abstract}
We present Fikiria, a two-stage SFT+GRPO recipe for Swahili mathematical reasoning. Its reward multiplies a format gate by a weighted sum of answer accuracy and the ChrF\texttt{++} similarity of the Swahili reasoning to an in-language reference chain of thought, so that training targets the surface fidelity of the reasoning trace and not only the final answer. We train on a 2.3k-problem Swahili dataset created by translating GSM8K with GPT-4.1 and filtering with SSA-COMET. On AfriMGSM Swahili, Fikiria-AfriqueQwen-14B reaches 79.2\%. Under the same fixed prompt and strict answer-format scoring, GPT-4o reaches 68.0\%, partly because it often fails the required answer format; IrokoBench reports 78.8\% for GPT-4o using a best-per-model prompt, a figure numerically close to ours but not directly comparable. In a small human evaluation with co-author annotators, GPT-4o's outputs were preferred overall. ChrF\texttt{++} measures surface fidelity, not the validity of individual reasoning steps.
\end{abstract}

%%─────────────────────────────────────────────────────────────────
\section{Introduction}
%%─────────────────────────────────────────────────────────────────

% \begin{itemize}

  Recent mathematical reasoning models often produce correct answers,
  but their intermediate reasoning is expressed in English, making it
  inaccessible to many students. This creates a mismatch between
  correctness and usability: a correct answer without an understandable
  explanation does not support learning. We study the following problem:
  given a math question in Swahili, generate both (i) a correct answer
  and (ii) a step-by-step reasoning chain in Swahili that a student can
  follow. We define our target as native-language reasoning quality: reasoning that is well-formed, expressed in fluent target-language text, and paired with a correct final answer. Whether individual steps causally support the answer is not verified by our automated training signal, but is assessed qualitatively with human evaluation on a subset of outputs (Section \ref{sec:native_speaker_protocol}).

  Prior work on mathematical reasoning (e.g., TAPO
  \citep{huang2026tapo}, MathOctopus \citep{chen2023mathoctopus})
  optimizes for answer accuracy, often treating reasoning as an
  auxiliary signal. CURE-Med \citep{onyame2026curemed} extends this
  to multilingual medical reasoning and enforces language consistency
  only on the final answer. \citet{barua_long_2026} explored how different stages of model development affect long reasoning traces beyond English. We explicitly optimize for the quality and native-language fidelity of reasoning in the target language with gated structured rewards applied to a \texttt{<rationale>} block.

  Our approach optimizes the reasoning trace directly rather than treating it as an auxiliary signal. Methods that target answer accuracy alone (TAPO, MathOctopus) or enforce language consistency only on the final answer (CURE-Med) can give full reward to a correct answer reached through English-language reasoning that is unusable to a Swahili-speaking student. More recent work rewards the language of the reasoning itself \citep{zhang2025mthinker, hwang2025learngloballyspeaklocally}. Fikiria differs in its reward construction: a multiplicative format gate (tags, an equation in the rationale, minimum length) that zeroes the reward for malformed outputs, applied to two separate terms, a parameter-free ChrF\texttt{++} similarity of the Swahili rationale to an in-language reference chain of thought and exact-match accuracy. We train on a Swahili math dataset we construct and evaluate on human-translated AfriMGSM.

  Our contributions are as follows:
  \begin{enumerate}
    \item A reward for Swahili reasoning traces that combines a multiplicative format gate, a ChrF\texttt{++} term against an in-language reference chain of thought, and exact-match accuracy, together with a 2.3k-problem Swahili training set.
    \item A two-stage recipe in which cold-start SFT bootstraps output structure and Swahili instruction-following before GRPO. In ablations on Qwen3, SFT+GRPO outperforms GRPO alone by 6.4 points at 8B, 12.0 points at 14B, and 2.0 points (within noise) at 32B.

  \end{enumerate}
% \end{itemize}

\begin{figure}[t]
\centering
\includegraphics[width=\linewidth]{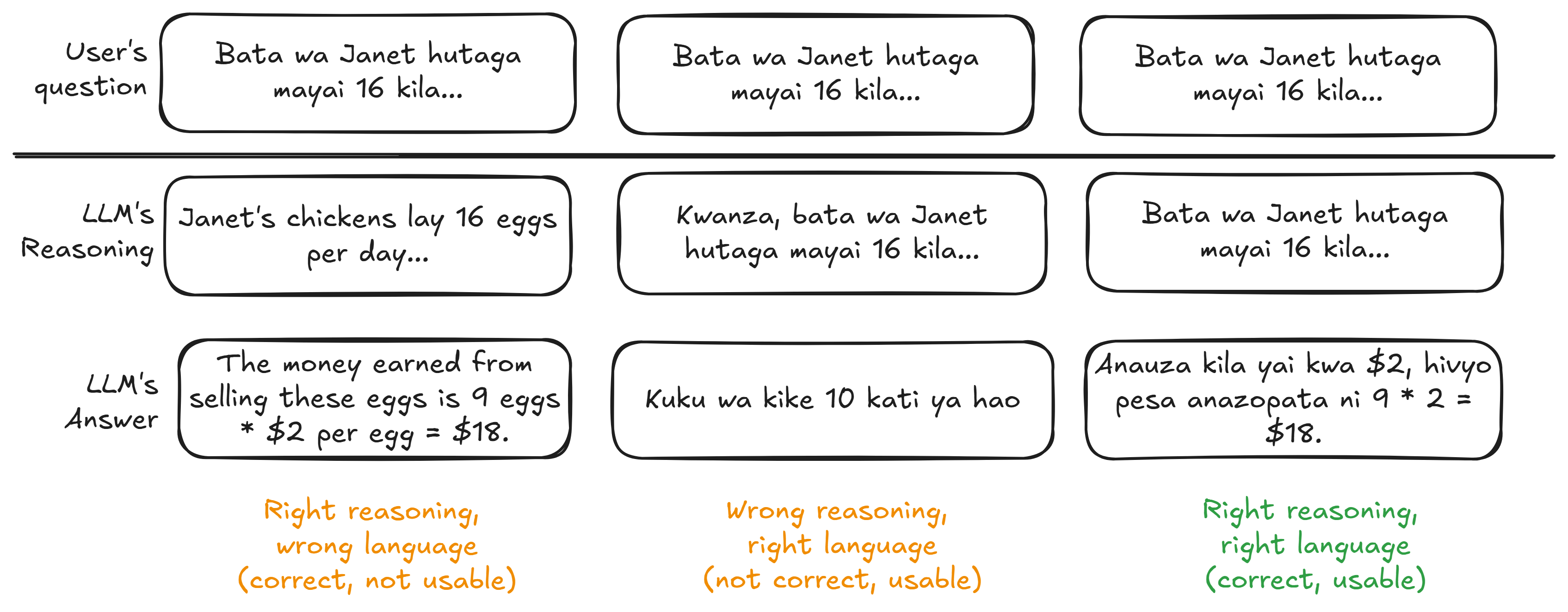}
\caption{Problem summary. Given a math question in Swahili, the LLM must generate both a correct answer and linguistically usable reasoning. Previous work has focused on improving accuracy while this paper considers both correctness of the final answer and usability of the reasoning trace through a two-stage SFT and GRPO pipeline described in Figure~\ref{fig:pipeline}.}
\label{fig:summary}
\end{figure}

\section{Related Work}
\label{sec:related}

Our work relates to three lines of research: accuracy-focused multilingual mathematical reasoning, rewards for language consistency in multilingual reasoning, and the quality of generated reasoning traces.

\paragraph{Multilingual mathematical reasoning.}
TAPO \citep{huang2026tapo} and MathOctopus \citep{chen2023mathoctopus} improve multilingual accuracy on GSM8K \citep{cobbe2021gsm8k} and MGSM \citep{shi2022language}. TAPO follows an understand-then-reason paradigm: the model translates the question into English before reasoning, and uses ChrF\texttt{++} \citep{popovic2017chrfpp} as a translation-quality reward. Its reasoning trace is therefore in English, whereas we target reasoning in the target language itself. MAPO \citep{she2024mapo} improves multilingual reasoning through alignment-as-preference optimization, and TRIT \citep{liu2026trit} jointly trains translation and reasoning in a self-improving loop that needs no external feedback. Neither rewards the language or quality of the reasoning trace itself.

\paragraph{Language-consistency rewards.}
Several recent methods use reinforcement learning with Group Relative Policy Optimization (GRPO; \citealp{shao2024grpo}) to keep reasoning in the input language. CURE-Med \citep{onyame2026curemed} applies GRPO to multilingual medical reasoning across 13 languages, including Swahili, with a linear reward combining correctness, language consistency and format ($\lambda=0.65$, $0.30$ and $0.05$). Language consistency is enforced on the final \texttt{<answer>}, not the reasoning trace, and the reward is not gated, so malformed outputs still receive partial credit. M-Thinker \citep{zhang2025mthinker} gates its reward on a rule-based Language Consistency (LC) check over the thinking and answer, and adds a Cross-lingual Thinking Alignment (CTA) reward computed by an external LLM judge against the model's own English reasoning. We avoid learned or judged rewards because of reward-hacking risk (Section~\ref{sec:GRPO}). \citet{hwang2025learngloballyspeaklocally} train Qwen2.5-7B-Instruct with a joint SFT and GRPO objective on machine-translated questions in ten MGSM languages, including Swahili, using a language-identification reward (and, in a later version, mT5 embedding-similarity rewards against the original English reasoning trace). They also introduce a factual-reasoning benchmark with Swahili. On MGSM Swahili their outputs stay in Swahili but math accuracy drops from 14.0\% to 2.1--3.7\%. \citet{qi-etal-2025-models} show that reasoning models often default to English or produce fragmented reasoning in other languages, and that forcing the reasoning language through prompting improves readability but lowers accuracy. This motivates optimizing the reasoning trace directly instead of relying on prompts.

\paragraph{Reasoning-trace quality.}
Supervising intermediate reasoning, rather than only final answers, has a long history in English-language settings \citep{bai_constitutional_2022, lightman_lets_2023}. \citet{barua_long_2026} study how different stages of model development affect long chain-of-thought (CoT) reasoning beyond English, focusing on accuracy. Surface metrics such as ChrF\texttt{++} measure text overlap with a reference rather than whether a reasoning trace causally supports the answer, a limitation we discuss in Section~\ref{sec:conclusion}.

\paragraph{Positioning.}
Gating a reward on format validity is not unique to our work: M-Thinker gates on format and language consistency, and TAPO masks its ChrF\texttt{++} translation reward with a format reward. Our gate checks only format (tags present, an equation in the rationale, minimum length). Language fidelity enters through a separate term, the ChrF\texttt{++} similarity of the Swahili rationale to an in-language reference chain of thought, alongside an exact-match accuracy term. Fikiria therefore differs from these methods in what the surface-metric reward is computed on and against what reference, and in being evaluated on human-translated AfriMGSM Swahili.

\paragraph{African NLP context.}
We build on the continued pre-training model family AfriqueLLM \citep{yu2026afriquellm} and the benchmark AfriMGSM \citep{ojewale2025afrimgsm}.

%%─────────────────────────────────────────────────────────────────
\section{Method}
%%─────────────────────────────────────────────────────────────────

\subsection{Problem formulation}

We consider the task of Swahili mathematical reasoning with native-language justification. Given a math problem posed in Swahili, the model must produce two outputs: a step-by-step reasoning trace in Swahili and a final numeric answer. We define success along two axes. A response is correct if the numeric answer matches the ground truth. A response has high native-language reasoning quality if the reasoning trace is expressed in fluent Swahili, contains mathematical steps consistent with a correct answer, and is legible to a native Swahili-speaking student. We measure this via surface-level linguistic fidelity (ChrF\texttt{++}) and answer correctness, and separately examine step-level causal support through human evaluation on a sample (Section \ref{sec:native_speaker_protocol}). Optimizing for correctness alone is insufficient: a model can produce the right answer via English reasoning, which fails the linguistic requirement even when the answer is right.

\subsection{Pipeline overview}

\begin{figure}[t]
\centering
\includegraphics[width=\linewidth]{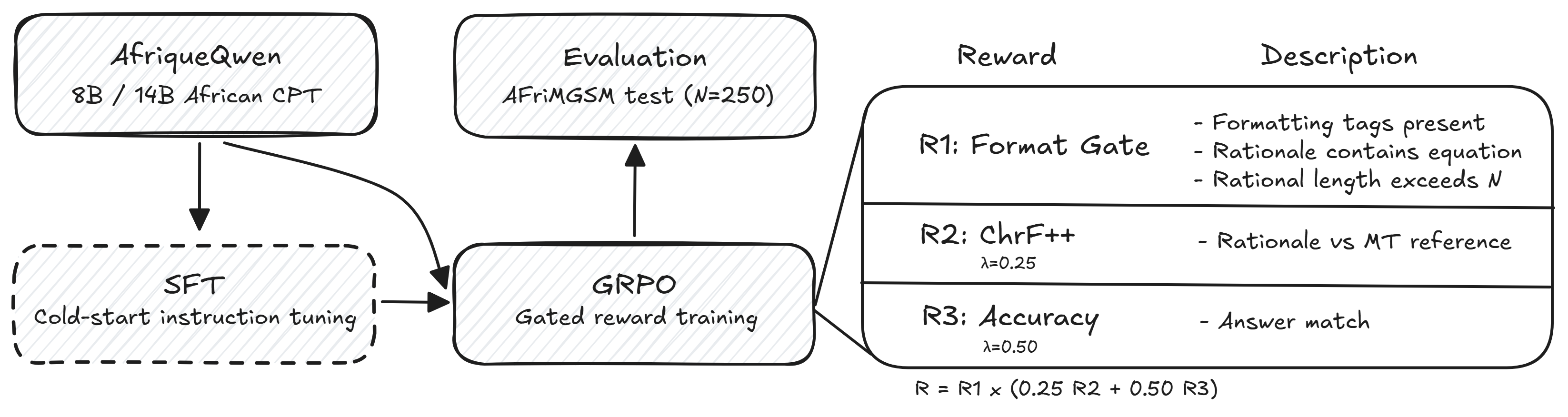}
\caption{Fikiria training pipeline. Starting from AfriqueQwen
(8B or 14B, African CPT from Qwen3), cold-start SFT teaches
Swahili instruction-following and the
\texttt{<rationale>}/\texttt{<answer>} output format.
GRPO then refines the model using a gated reward:
R$_1$ (format gate) multiplicatively gates
R$_2$ (ChrF\texttt{++} of \texttt{<rationale>} against a
Swahili MT reference, $\lambda{=}0.25$) and
R$_3$ (exact-match accuracy on \texttt{<answer>},
$\lambda{=}0.50$).
The dashed SFT border and bypass arrow indicate that
GRPO-only ablation variants omitting SFT are also evaluated
(Table~\ref{tab:ablation-progression}).}
\label{fig:pipeline}
\end{figure}

Figure~\ref{fig:pipeline} shows the summary of the two-stage, SFT+GRPO, training along with additional details on the GRPO reward construction.

\subsection{Dataset construction}
\label{sec:dataset_construction}

The Grade School Math 8K dataset (GSM8K) contains 8.5k grade school math question-answer pairs and is a popular English-only benchmark for math reasoning \citep{cobbe2021gsm8k}. Several datasets, such as MGSM \citep{shi2022language} and AfriMGSM \citep{ojewale2025afrimgsm}, extend GSM8K to non-English languages, including Swahili, via machine translation or human translation. However, these are too small for our purposes ($N$~=~$250$).

To construct our dataset, we translated all question-answer pairs into Swahili using GPT-4.1 with a 3-shot prompt drawn from AfriMGSM (see Appendix~\ref{app:translation_templates} for templates). We then assessed translation quality using SSA-COMET-MTL \citep{li_ssa-comet_2025}, a reference-free quality estimator (QE), and conducted a human spot check (see Figure~\ref{fig:ssa_comet_qe_distribution} in Appendix~\ref{app:data}). For the remainder of this work, we retain only entries whose translated question and answer both score above 0.65 on SSA-COMET-MTL, reducing the train and test splits to roughly 2.3k and 400 entries, respectively.

The filtering threshold was chosen to ensure high translation fidelity, which can significantly impact downstream applications \citep{peter_mind_2025}. We adopted the value recommended by the metric authors [D. Adelani, private comm.] and validated it through human spot-checks and cross-dataset comparison.

For the human spot-check, we randomly sampled 20 examples distributed around the threshold and asked two native Swahili speakers to annotate each translation for grammatical and logical consistency (see Section~\ref{sec:native_speaker_protocol} for annotator selection). The small sample size limits statistical power, but the human scores are well-balanced around the metric as a linear predictor: roughly half fall below the value predicted by linear regression at the 0.65 threshold (see Figure~\ref{fig:ssa_comet_qe_spot_check} in Appendix~\ref{app:data}).

In addition to the spot check, we compared SSA-COMET-MTL metrics for our dataset to  AfriMGSM, which is an extension of MGSM \citep{shi2022language, ojewale2025afrimgsm}. The GPT-4.1 translations score similarly for values greater than 0.65 while for samples with lower scores GPT-4.1 tends to score high (see Figure~\ref{fig:dataset_comet_compare}). Across all three reference datasets, for samples exceeding this threshold, at least 89\% of GPT-4.1 samples also exceed the threshold. This indicates over all quality and agreement between the machine-translated entries to current standard datasets.
% https://huggingface.co/spaces/gatesfoundation/kr-jupyter

We release the translations and corresponding SSA-COMET-MTL scores on Hugging Face. At roughly 2.3k training and 400 test entries after filtering, this dataset is substantially larger than MGSM and AfriMGSM, both of which contain 250 entries for testing and 8 for training.

\subsection{Supervised fine-tuning}

The first stage of our training pipeline is a supervised fine-tuning round that jointly entrains the chat template, Swahili response, and reasoning style. We train for completion over 3 epochs of the combined Swahili question and COT dataset using TRL to minimize the standard token-level cross-entropy loss \citep{vonwerra2020trl}.

\subsection{GRPO with native-language rewards}
\label{sec:GRPO}

Figure~\ref{fig:pipeline} shows the reward structure and equation, including R$_1$ (format gate) that gates all rewards, R$_2$ (ChrF\texttt{++} vs Swahili machine-translated reference):Swahili--Swahili, and R$_3$ (exact match from \texttt{<answer>}).  

The format gating requires that the generated content be of sufficient length, contain tags for the rationale and answer, and the rationale must contain an equation.  By gating the reward function with the format, we enforce the answer can be parsed and includes a math equation in the rationale, which we found to be the case for all languages in the AfriMGSM dataset train split. See Appendix~\ref{app:evaluation_templates} for the full prompt template.

For the language quality reward, R$_2$, we require a reward that incentivizes Swahili output while being robust to reward hacking. We considered SSA-COMET-MTL \citep{li_ssa-comet_2025}, an embedding-based quality estimator well-suited for African languages, but followed \citet{huang2026tapo} in avoiding learned embedding-based rewards, which are susceptible to reward hacking via adversarial outputs that score highly without genuine quality improvement \citep{skalse_defining_2025}. A second reason to avoid a learned metric here is signal separation: R$_2$. should measure language fidelity independently of answer correctness, which is already captured by R$_3$. Conflating these into a single learned metric would obscure whether the model is being rewarded for reasoning in Swahili or for getting the answer right. Instead, we use ChrF\texttt{++} \citep{popovic2017chrfpp}, a surface-level character n-gram metric with no learned parameters. To validate that ChrF\texttt{++} serves as an effective proxy for Swahili language fidelity, we examined GlotLID Swahili confidence scores \citep{kargaran_glotlid_2023} during initial experiments and found that optimizing ChrF\texttt{++} against a Swahili--Swahili reference produced a steady increase in GlotLID Swahili confidence over training steps. This indicates that surface-level character overlap with a Swahili reference is a proxy for target-language output.

\subsection{Human evaluation protocol}
\label{sec:native_speaker_protocol}

We conduct a blind pairwise human evaluation to compare Fikiria-AfriqueQwen-14B against GPT-4o on Swahili mathematical reasoning. For each example, annotators are shown a Swahili math question together with two anonymized model outputs presented in randomized order. Each output consists of a step-by-step solution in Swahili and a final numeric answer. To minimize stylistic bias unrelated to reasoning quality, we normalize both outputs by removing model-specific formatting (e.g., XML-style tags) while preserving the underlying reasoning content and final answer.

We use two native Swahili-speaking researchers (co-authors) with STEM backgrounds as annotators. This choice reflects the practical constraint that evaluating mathematical reasoning in a low-resource language requires both domain knowledge and high language proficiency, which are difficult to obtain simultaneously. All evaluations are conducted in a fully blind pairwise setting with randomized output ordering and normalized formatting to remove model-specific cues. While using co-authors as annotators may introduce bias, we treat this evaluation as a qualitative assessment of reasoning quality rather than a large-scale benchmark. The annotators are based in different regions (Kenya and Tanzania), reflecting dialectal and stylistic variation in Swahili, which introduces additional subjectivity but better captures real-world variation in language use. This setup enables fine-grained assessment of reasoning faithfulness that would be difficult to obtain with non-expert annotators. Each example is independently annotated by both annotators. See Appendix~\ref{app:evaluation_example} for a screenshot of the data collection interface.

We report inter-annotator agreement alongside randomized, pairwise preference results. We emphasize that this human evaluation is intended as a qualitative complement to automatic metrics, rather than a standalone large-scale benchmark.

Annotators are asked to select which output is better overall, with three possible labels: \textit{A}, \textit{B}, or \textit{Tie}. They are instructed to base this decision on four criteria: (i) correctness of the final answer, (ii) whether the intermediate reasoning steps causally support the answer, (iii) clarity and usefulness for a Swahili-speaking student, and (iv) naturalness of the Swahili, including avoidance of code-switching or unnatural phrasing. This protocol is designed to better match our target notion of \emph{faithful native-language reasoning} than scalar fluency judgments alone. 

After making the pairwise preference decision, annotators assign a single categorical tag indicating the primary reason for their choice. The tag set is: \textit{correctness}, \textit{reasoning quality}, \textit{clarity/helpfulness}, \textit{language quality}, or \textit{other}. These tags allow us to distinguish whether preferences are driven by mathematical validity, reasoning faithfulness, pedagogical usefulness, or linguistic quality, rather than by overall style alone. The annotator is not required to provide a reason for a tie.

Before the main evaluation, annotators complete a short calibration phase on a separate set of examples in order to align interpretation of the criteria, especially the distinction between fluent Swahili and reasoning that genuinely supports the final answer. We then report pairwise win/loss/tie rates, category-level preference breakdowns, and inter-annotator agreement. In addition, we separately summarize common error types observed during annotation, including code-switching, fabricated intermediate steps, and correct final answers supported by incomplete or misleading reasoning. 

%%─────────────────────────────────────────────────────────────────
\section{Experiments}
%%─────────────────────────────────────────────────────────────────

\subsection{Experimental setup}

\begin{figure}[t]
\centering
\begin{minipage}{0.48\linewidth}
  \includegraphics[width=\linewidth]{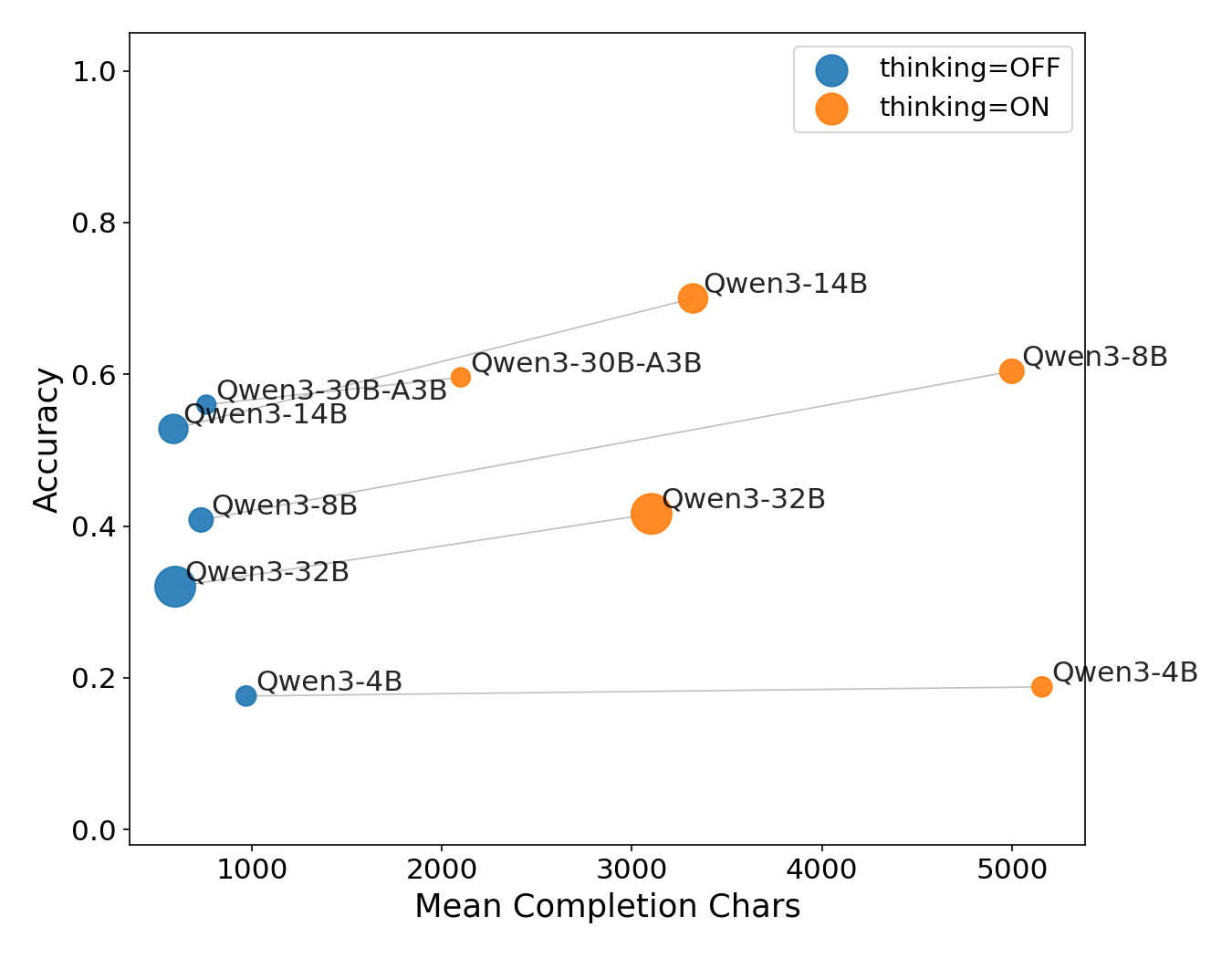}
\end{minipage}\hfill
\begin{minipage}{0.48\linewidth}
  \includegraphics[width=\linewidth]{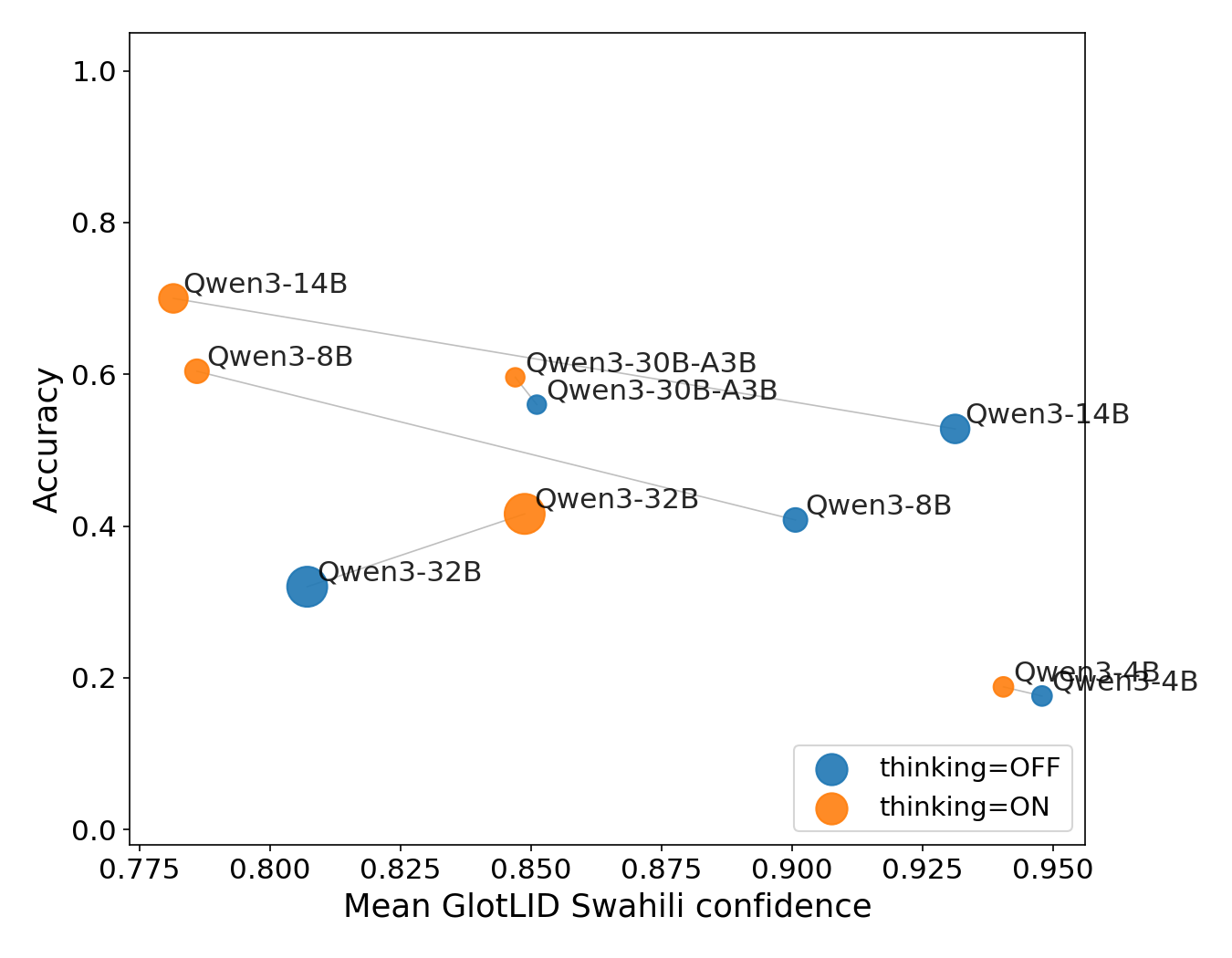}
\end{minipage}
\caption{Effect of \texttt{thinking=ON} vs \texttt{thinking=OFF} on
Qwen3-family models (AfriMGSM Swahili test, 250 examples).
Marker size is proportional to active parameter count.
Grey lines connect the same model under both conditions. Left: Accuracy vs.\ mean completion length (characters).
\texttt{thinking=ON} yields marginally higher accuracy at
3--6$\times$ greater completion cost.
Right: Accuracy vs.\ mean GlotLID Swahili confidence. 
\texttt{thinking=OFF} produces higher language identification confidence, an indicator for increased fidelity.
We use \texttt{thinking=OFF} throughout to increase native language performance and reduce compute.}
\label{fig:thinking-tradeoff}
\end{figure}

Qwen3 models support an extended reasoning mechanism
(\texttt{enable\_thinking=ON}) that generates a native
\texttt{<think>} block before the final response.
Figure~\ref{fig:thinking-tradeoff} shows the accuracy--efficiency tradeoff across Qwen3-family models on AfriMGSM Swahili. While \texttt{thinking=ON} yields marginally higher accuracy, it does so at 3--6$\times$ greater completion length with no consistent gain in Swahili language fidelity, and in several models reduces it.  This is likely because extended \texttt{<think>} blocks
introduce English reasoning that lowers GlotLID Swahili confidence. Since our GRPO training uses \texttt{<rationale>} tags to supervise reasoning directly, the native thinking mechanism is redundant and
conflicts with our output format. We therefore use \texttt{enable\_thinking=False} throughout all SFT, GRPO, and evaluation runs.

Figure~\ref{fig:thinking-tradeoff} and Table~\ref{tab:ablation-progression} additionally show that accuracy is not tied to parameter count within the Qwen3 family. Qwen3-32B scores 32.0\% zero-shot, below the smaller Qwen3-8B (40.8\%) and Qwen3-14B (52.8\%). One possible explanation, which we did not test, is that larger models trained predominantly on high-resource languages have stronger English-reasoning priors, while smaller models may be more flexible in following Swahili-language instructions. After GRPO alone, performance of Qwen3-32B is greater than the smaller models (68.8\% vs.\ 53.6\% for Qwen3-14B and 51.2\% for Qwen3-8B), suggesting that task-specific optimization helps surface multilingual capability.

We focus on AfriqueQwen-8B and AfriqueQwen-14B
\citep{yu2026afriquellm} as our primary models. These models are part of a model family that has been adapted to 20 African languages from Qwen3-8B, Qwen3-14B (respectively) through continued pre-training and have good baseline performance in both Swahili and reasoning. We also compare the performance of the base models as well as larger variants in the same model family: Qwen3-32B, and Qwen3-30B-A3B (instruct, zero-shot). Finally, we also consider metrics from TAPO-ChrF\texttt{++} \citep{huang2026tapo}, Qwen2.5-3B, Llama3.2-3B, and IrokoBench \citep{ojewale2025afrimgsm} (which includes frontier models). All local inference uses \texttt{enable\_thinking=False}.

SFT and GRPO training were conducted on a cloud instance with 8$\times$A100 80GB GPUs (96 vCPU, 1136GB RAM, 640GB VRAM) and 4$\times$A100 instance
(48 vCPU, 568GB RAM, 320GB VRAM). Total wall-clock training time ranged from 7--15 hours per model for GRPO stage.  GRPO training used group size $G$~=~$8$, temperature $T$~=~$1.2$, maximum completion length 512 tokens, and learning rate $2\times10^{-6}$.
Models were fine-tuned with LoRA \citep{hu2021loralowrankadaptationlarge} (rank 32, $\alpha$~=~$64$, dropout 0.05) applied to query, key, value, output, gate, up, and down projection layers.
Effective batch size was 48 prompts per update step
(4 per device $\times$ 4 gradient accumulation steps $\times$ 3 processes), each generating 8 completions, for 384 total rollouts per step. Reward weights were $\lambda_{\mathrm{chrf}}$~=~$0.25$ and
$\lambda_{\mathrm{acc}}{=}0.50$ with no KL penalty
($\beta{=}0$, DAPO asymmetric clipping \citep{yu_dapo_2025}).  Inference used greedy decoding ($T{=}0$) with a maximum of
512 tokens (\texttt{enable\_thinking=False} and 4096-tokens for \texttt{thinking=ON} for results shown in Figure~\ref{fig:thinking-tradeoff}.

\subsection{Main results}

\begin{figure}[t]
\centering
\includegraphics[width=\linewidth]{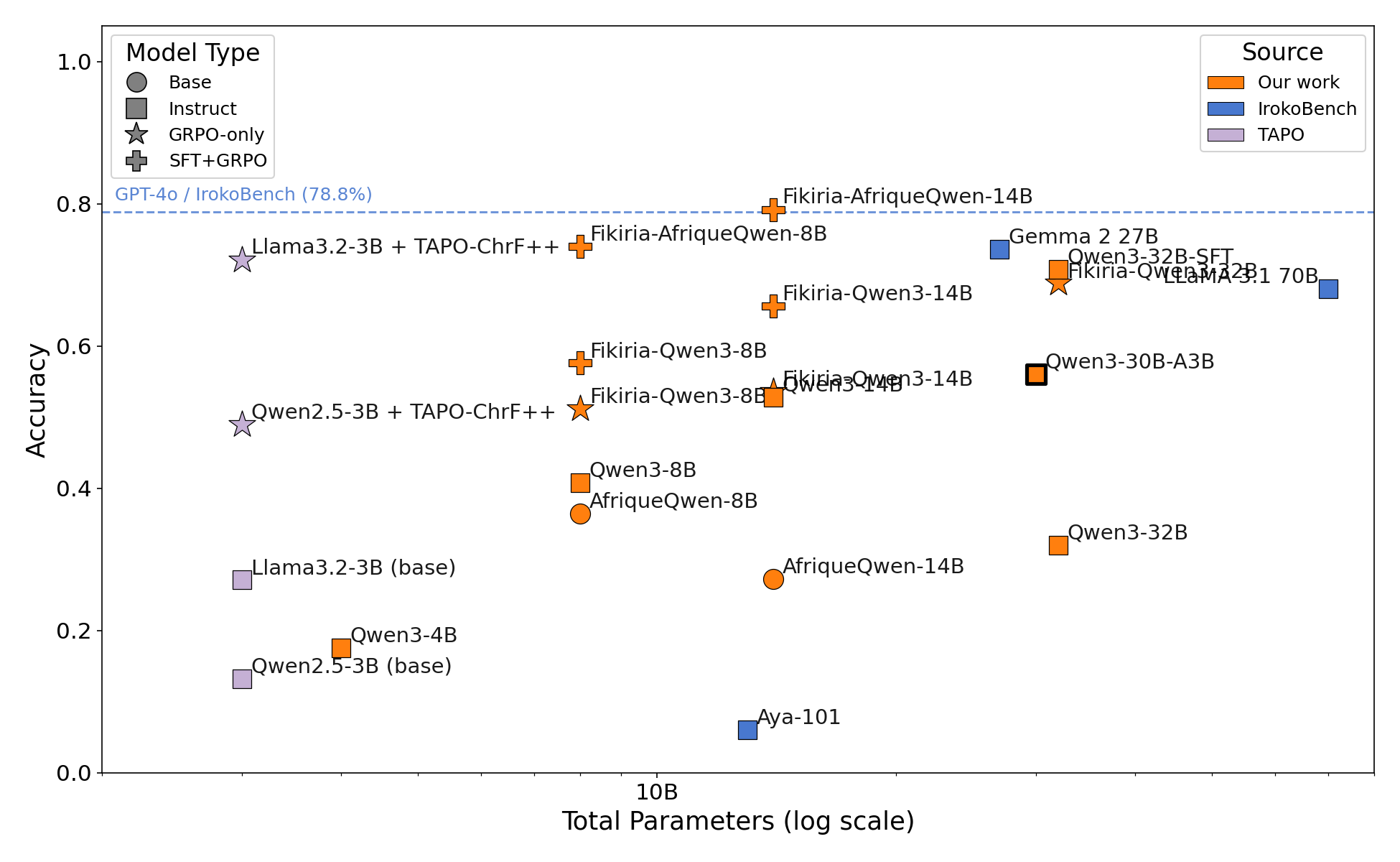}
\caption{Swahili math accuracy (AfriMGSM test, 250 examples) vs.\
total parameter count (log scale).
Marker shape: circle = base, square = instruct, star = GRPO-trained.
Thick edges = MoE (total parameters $\gg$ active parameters during
inference). Color: orange = ours, blue = IrokoBench
\citep{ojewale2025afrimgsm}, purple = TAPO \citep{huang2026tapo}.
GPT-4o shown as blue dashed reference line (IrokoBench best-per-model prompt; parameter count undisclosed).}
\label{fig:size-vs-accuracy}
\end{figure}

Figure~\ref{fig:size-vs-accuracy} plots AfriMGSM Swahili accuracy against total parameter count. Fikiria-AfriqueQwen-14B reaches 79.2\% ($\pm$2.6). Under the same prompt, decoding and strict scorer, GPT-4o reaches 68.0\% ($\pm$3.0), partly because it often wraps answers in extra characters (for example \texttt{<answer>\$18</answer>}), which the strict scorer marks wrong. The 78.8\% reported for GPT-4o by \citet{ojewale2025afrimgsm} used best-performing prompts per model and is not directly comparable. Fikiria-AfriqueQwen-14B also exceeds Gemma~2~27B (73.6\%). Table~\ref{tab:ablation-progression} shows this result in the context of training stages and baselines. Zero-shot, AfriqueQwen-14B scores 27.2\%, while Qwen3-14B instruct reaches 52.8\% without any task-specific training. After SFT$\to$GRPO, Fikiria-AfriqueQwen-14B gains +52.0 points over its own base and +26.4 points over the Qwen3-14B instruct baseline. Under the same SFT$\to$GRPO pipeline, the AfriqueQwen backbones outperform their Qwen3 counterparts by +13.6 points at 14B (79.2\% vs.\ 65.6\%) and +16.4 points at 8B (74.0\% vs.\ 57.6\%), suggesting that African continued pre-training provides a stronger foundation for GRPO than the original instruction-tuned model. Fikiria-AfriqueQwen-8B reaches 74.0\%, +37.6 points over its base.

Fikiria-AfriqueQwen-14B (79.2\%) scores higher than TAPO-ChrF\texttt{++} on Llama3.2-3B (72.1\%), which targets English reasoning and is evaluated on MGSM Swahili rather than AfriMGSM. The two results differ in benchmark, model size and backbone, so we do not attribute the difference to any single factor.

IrokoBench models \citep{ojewale2025afrimgsm} were evaluated with the best-performing prompt template per model; our models use a fixed
\texttt{<rationale>}/\texttt{<answer>} prompt consistent with GRPO
training, and prompt choice can meaningfully affect accuracy on
AfriMGSM. TAPO \citep{huang2026tapo} reports on MGSM Swahili rather
than AfriMGSM; these points situate the training paradigm but are not
directly comparable. Mixture-of-Experts (MoE; thick edges) are plotted at total
parameter count; effective compute is closer to active parameter count
(e.g.\ Qwen3-30B-A3B uses ${\sim}$3B active parameters per forward
pass). Aya-101 (13B, mT5-XXL) scores 6\%, consistent with its
multilingual generalist design \citep{ojewale2025afrimgsm}.

\subsection{Ablation Studies}
\label{sec:ablations}

Table~\ref{tab:ablation-progression} shows GRPO-only and SFT+GRPO training on three Qwen3 models (single training run per cell). These ablations isolate the effect of cold-start SFT; they do not isolate the reward design (see limitations in Section~\ref{sec:conclusion}), and we did not run a GRPO-only ablation on AfriqueQwen.

SFT adds the most at 8B and 14B. Without it, GRPO must simultaneously discover the output format, Swahili instruction following, and reasoning quality, which may be harder at smaller scale. Qwen3-8B reaches 51.2\% with GRPO alone (+10.4 points over the 40.8\% instruct baseline) and 57.6\% with SFT+GRPO (+16.8 points); the 6.4-point difference between the two is modest relative to the standard errors. The effect is clearer at 14B: GRPO alone improves on the 52.8\% baseline by only 0.8 points (53.6\%, within noise), while SFT+GRPO reaches 65.6\% (+12.8 points, +12.0 over GRPO-only).

At 32B, GRPO alone accounts for nearly all of the improvement: Qwen3-32B reaches 68.8\% (+36.8 points over its 32.0\% baseline), and SFT adds only 2.0 points (70.8\%), within one standard error. One possible explanation, which we did not test, is that the larger instruction-tuned model can acquire the output format during GRPO without cold-start SFT, whereas the smaller models benefit from SFT first.

\subsection{Human evaluation}
\label{sec:human_eval}

Following the protocol in Section~\ref{sec:native_speaker_protocol}, we collected $N$~=~54 samples rated by both annotators (108 entries total). Assuming a Dirichlet(1,1,1) prior on ($P_A,P_B,P_\mathrm{Tie}$), the posterior probability that GPT-4o's true preference rate exceeds Fikiria's is 0.999, with an estimated gap of 24.5 points (95\% credible interval: +9 to +40). Inter-rater agreement was 41\%, and the two annotators used the "Tie" label very differently (51\% vs. 0\%). The direction of the preference is therefore robust, but the magnitude should be treated as approximate. 

Notably, Fikiria-AfriqueQwen-14B was judged as good as or better than GPT-4o on 30--60\% of samples (depending on annotator) and outright preferred 11--30\% of the time. The reasons given for each preference also differed significantly ($\chi^2$~=~18.89, dof~=~4, $p$~=~$8\times10^{-4}$): Fikiria-AfriqueQwen-14B was preferred for better Swahili (12 out of 22 cases), while GPT-4o preferences split between clearer explanation and better reasoning (with better Swahili cited in only 10\% of cases). Although GPT-4o is preferred overall, Fikiria-AfriqueQwen-14B was preferred most often for naturalness of Swahili, suggesting that its main advantage over GPT-4o lies in language quality rather than reasoning clarity. Given the small sample, the co-author annotators, and moderate agreement (41\%), we treat these results as suggestive.

\begin{table}[t]
\small
\centering
\renewcommand{\arraystretch}{1.3}
\begin{tabular}{@{}llrrrr@{}}
\toprule
& & \multicolumn{4}{c}{\textbf{Training Stage}} \\
\cmidrule(lr){3-6}
\textbf{Model} & \textbf{Size} & \textbf{Base} & \textbf{Instruct (baseline)} & \textbf{GRPO} & \textbf{Final (SFT+GRPO)} \\
\midrule

\addlinespace[2pt]
\multicolumn{6}{@{}l}{\small\textit{AfriqueQwen (African CPT) --- AfriMGSM test set}} \\
AfriqueQwen & 8B & \textbf{36.4 (±3.0)} &  &  & 74.0 (±2.8) \\
AfriqueQwen & 14B & 27.2 (±2.8) &  &  & \textbf{79.2 (±2.6)} \\

\midrule
\addlinespace[2pt]
\multicolumn{6}{@{}l}{\small\textit{Qwen3 --- AfriMGSM test set}} \\
Qwen3 & 4B &  & 17.6 (±2.4) &  &  \\
Qwen3 & 8B &  & 40.8 (±3.1) & 51.2 (±3.2) & 57.6 (±3.1) \\
Qwen3 & 14B &  & 52.8 (±3.2) & 53.6 (±3.2) & 65.6 (±3.0) \\
Qwen3 & 32B &  & 32.0 (±3.0) & \textbf{68.8 (±2.9)} & 70.8 (±2.9) \\
Qwen3 & 30B-A3B (MoE) &  & \textbf{56.0 (±3.1)} &  &  \\

\midrule
\addlinespace[2pt]
\multicolumn{6}{@{}l}{\small\textit{Reference model, our run (same prompt, decoding and scorer as our models)}} \\
GPT-4o (our prompt) & & & 68.0 (±3.0) & & \\

\midrule
\midrule
\addlinespace[2pt]
\multicolumn{6}{@{}l}{\small\textit{Prior work: IrokoBench \citep{ojewale2025afrimgsm} on AfriMGSM test set$^{\ddagger}$}} \\
\rowcolor{gray!8}Aya-101   & 13B &  &  6.0 &  &  \\
\rowcolor{gray!8}Gemma 2   & 27B &  & 73.6 &  &  \\
\rowcolor{gray!8}Llama 3.1 & 70B &  & 68.0 &  &  \\
\rowcolor{gray!8}GPT-4o    &     &  & 78.8 &  &  \\

\midrule
\addlinespace[2pt]
\multicolumn{6}{@{}l}{\small\textit{Prior work: TAPO \citep{huang2026tapo} on MGSM Swahili test set$^{\dagger}$}} \\
\rowcolor{gray!8}Qwen2.5  & 3B &  & 13.2 & 49.0 &  \\
\rowcolor{gray!8}Llama3.2 & 3B &  & 27.1 & 72.1 &  \\

\bottomrule
\end{tabular}

\caption{\small
Swahili math accuracy (\%) and standard error by model and training stage.
AfriqueQwen, Qwen3, and IrokoBench results are reported on the AfriMGSM test set (250 examples). $^{\ddagger}$IrokoBench values are taken from \citet{ojewale2025afrimgsm}. They use the best-performing prompt per model and have no reported standard error; our models use a single fixed \texttt{<rationale>}/\texttt{<answer>} prompt.
$^{\dagger}$TAPO values are taken from \citet{huang2026tapo}, have no standard error here, and are measured on MGSM Swahili, so they are not directly comparable.
GPT-4o (our prompt) is a single run at temperature 0 with the same prompt, decoding and strict scorer as our models; the IrokoBench GPT-4o value used a best-per-model prompt.
}
\label{tab:ablation-progression}
\end{table}

\section{Conclusion}
\label{sec:conclusion}
We present Fikiria, a two-stage SFT+GRPO recipe for Swahili mathematical reasoning whose reward multiplies a format gate by a ChrF++ term on the Swahili rationale and an exact-match accuracy term. Fikiria-AfriqueQwen-14B achieves 79.2\% on AfriMGSM Swahili versus 68.0\% for GPT-4o under the same prompt and strict scorer, while a small human evaluation favors GPT-4o's reasoning overall. Its accuracy is also higher than the 72.1\% reported for TAPO-ChrF++, though this value is measured on MGSM Swahili and is not directly comparable. Our results are consistent with the view that a surface-level reward on the reasoning trace can be optimized alongside accuracy, and that African CPT provides a stronger foundation in our setting than instruction tuning alone. A limited human evaluation further examines whether reasoning steps causally support the final answer, though on a small, moderately-agreeing sample (Section \ref{sec:human_eval}). Limitations include the small AfriMGSM evaluation set (250 examples), and  restriction to a single language and task. ChrF\texttt{++} is a surface-level metric of linguistic overlap with a reference translation; it does not verify that individual reasoning steps causally support the final answer, so our automated reward signal should be understood as targeting native-language fluency rather than step-level logical validity. Our experiments also do not ablate the reward design (ChrF\texttt{++} versus accuracy-only, or the format gate versus a plain weighted sum), so the contribution of each reward component is not isolated; all trained results are from a single training run per configuration; we did not run a GRPO-only ablation on AfriqueQwen; and because AfriMGSM is derived from GSM8K, which is possibly present in the pre-training data of the base models, some of the reported accuracy may reflect contamination rather than reasoning. Causal support between reasoning steps and the final answer is instead assessed qualitatively by human annotators on a small sample (N=54, Section \ref{sec:human_eval}); inter-annotator agreement on this sample was moderate (41\%), so these judgments should be treated as suggestive rather than a validated measure of faithfulness at scale. Our training data is constructed with machine translation, which may introduce systematic artifacts, mistranslations near the quality threshold, and biases inherited from the source model (GPT-4.1); downstream effects on reasoning quality are difficult to fully characterize. Human evaluation is small-scale and inherently subjective, with annotator disagreement reflecting both dialectal variation and the difficulty of judging reasoning quality in a low-resource language. Future work includes extending to other African languages, applying the gated reward framework to other low-resource reasoning tasks, developing automated metrics that more directly assess step-level causal validity, and scaling the human evaluation protocol.

\section{Acknowledgments and Funding}
This work is based on research conducted by Institute for Disease Modeling, a research group within, and solely funded by, the Gates Foundation. The funders had no role in the analysis, or interpretation of the data; the preparation, review, or approval of the report; or the decision to submit the manuscript for publication. Some authors are employees of the Gates Foundation, however, this study does not necessarily represent the views of the Gates Foundation.

\bibliography{references}

\appendix

\clearpage

\section{Dataset statistics and SSA-COMET-MTL score distribution}
\label{app:data}

\begin{figure}[h]
\includegraphics[width=\textwidth]{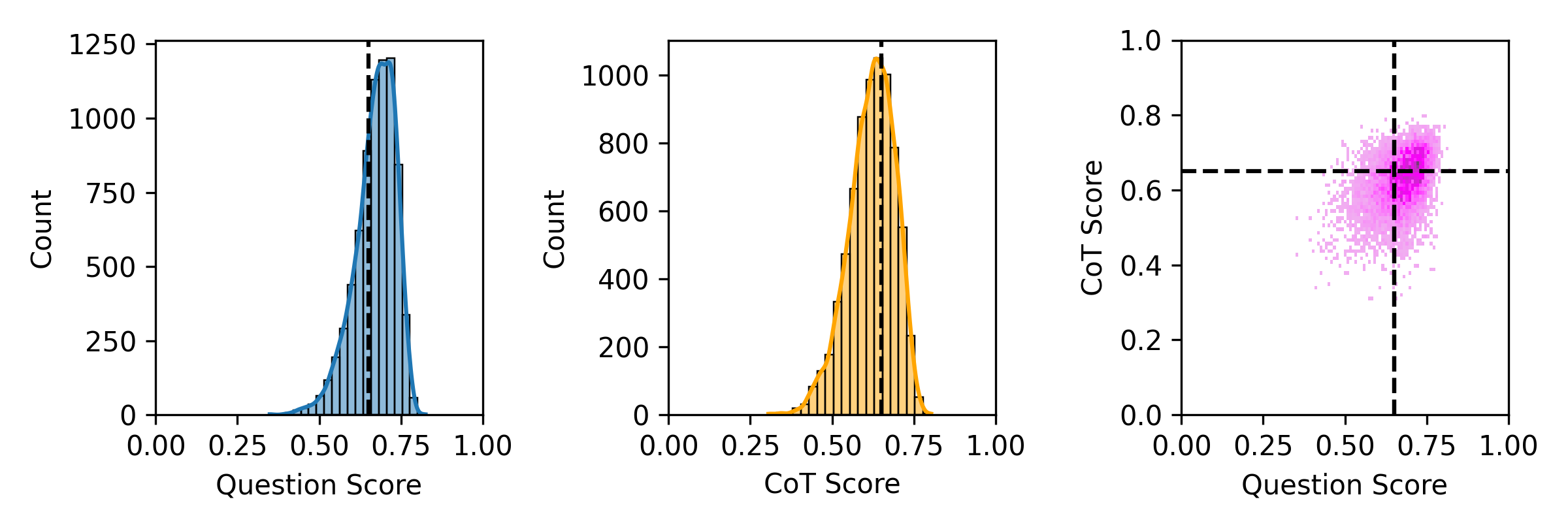}
\caption{Individual distributions of SSA-COMET-MTL QE scores for GPT-4.1 translated questions (left) and chain of thought (center) along with the joint distribution (right). Selection threshold of 0.65 is shown as dashed lines.}
\label{fig:ssa_comet_qe_distribution}
% https://github.com/krosenfeld/multilingual_math_reasoning/blob/main/analysis/gsm8k_comet_figures.py
\end{figure}

Figure~\ref{fig:ssa_comet_qe_distribution} shows SSA-COMET-MTL QE scores for GPT-4.1 (gpt-4.1-2025-04-14) translations of GSM8K English-to-Swahili questions and CoT for the train split. Distributions between the train and test splits were indistinguishable to a KS test. The distributions between translated questions and CoT are significantly different ($p$~$\ll$~$0.05$, KS-statistic~=~$0.31$). Filtering for scores greater than 0.65 reduces the test split to 416 (out of 1319) entries and the train split to 2308 (out of 7473).

\begin{figure}[h]
\begin{center}
    
\includegraphics[width=0.6\textwidth]{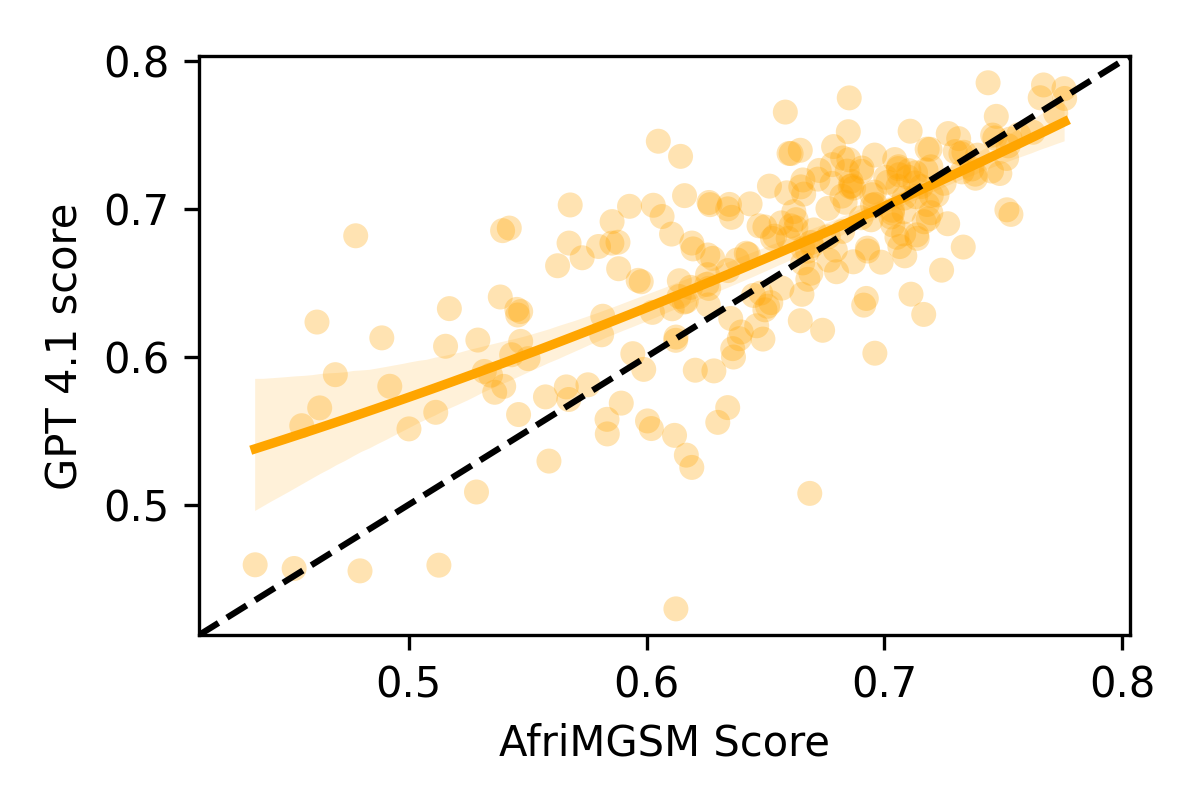}
\caption{Comparison of SSA-COMET-MTL QE scores for AfriMGSM (yellow) questions on a per sample basis to the GPT-4.1 translation presented in this work.}
\label{fig:dataset_comet_compare}
\end{center}
\end{figure}

Figure~\ref{fig:dataset_comet_compare} compares the SSA-COMET-MTL scores for GPT-4.1 translations compared to the AfriMGSM test split and a second-order polynomial regression \citep{Waskom2021}. GPT-4.1 tends to score higher than AfriMGSM, particularly for entries with lower scores.

\begin{figure}[h]
\includegraphics[width=\textwidth]{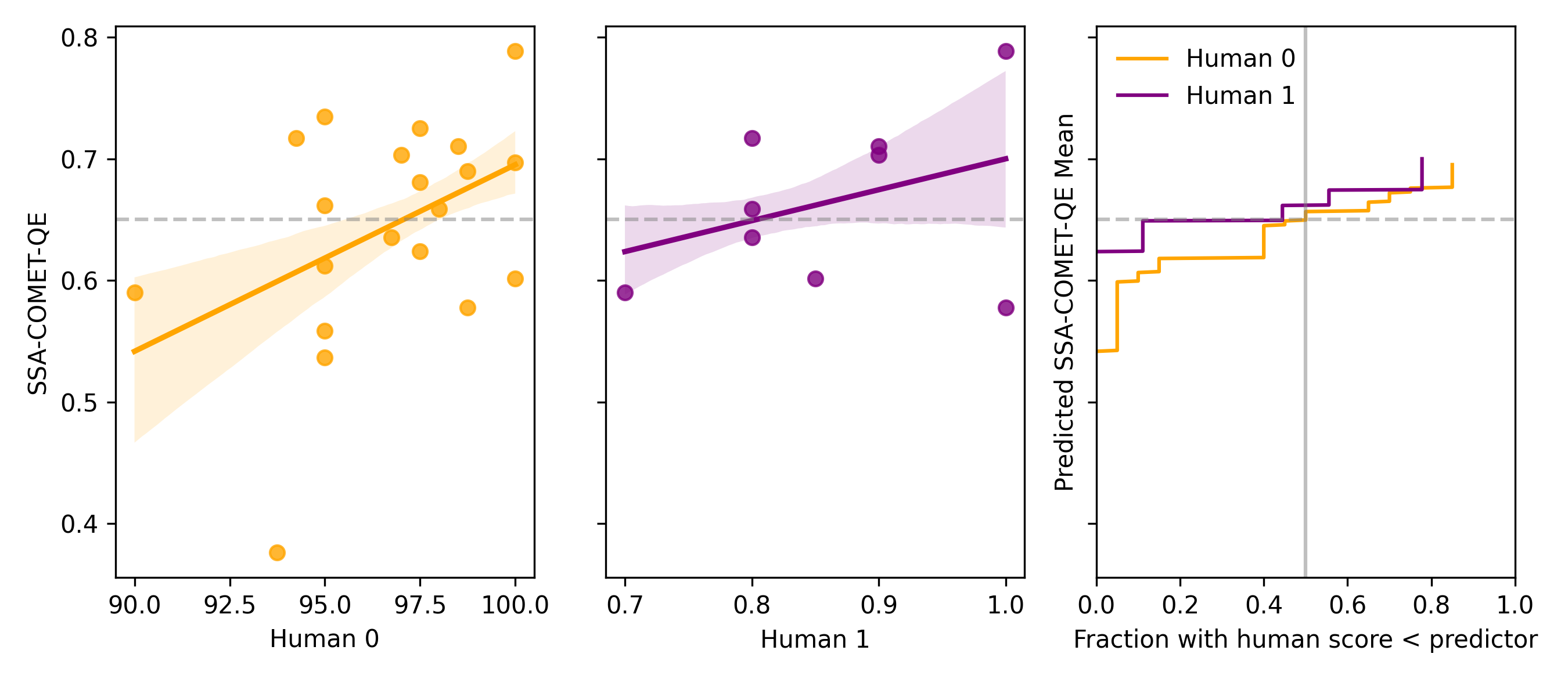}
\caption{Human spot check of GSM8K GPT-4.1 translations compared to the SSA-COMET-MTL metric (average of the Question and CoT).}
\label{fig:ssa_comet_qe_spot_check}
\end{figure}

Figure~\ref{fig:ssa_comet_qe_spot_check} shows the results of the human spot check of the quality of the generated text. Both annotators were given the same set of 20 question / answer pairs of GPT-4.1 translations to evaluate. Completion rates were 20/20 and 9/20. The annotators were asked to score the translation on a range of 0-100 but the resulting scalings were quite different (90 to 100 vs 70 to 100). However, when we look at the overall distribution of human scores to the SSA-COMET-MTL score we see a broad pattern of agreement without threshold. This is indicated by how a simple linear regression on the scores intersects the 0.65 threshold at a point that bisects both samples. 

Taken together, this spot check corroborates the choice of 0.65 as a reasonable delineation for high-quality translation although the signal is still noisy. We prefer to err on the side of high-quality since errors will negatively impact our training objectives.

\section{Human evaluation example}
\label{app:evaluation_example}

\begin{figure}[h]
    \includegraphics[width=\textwidth]{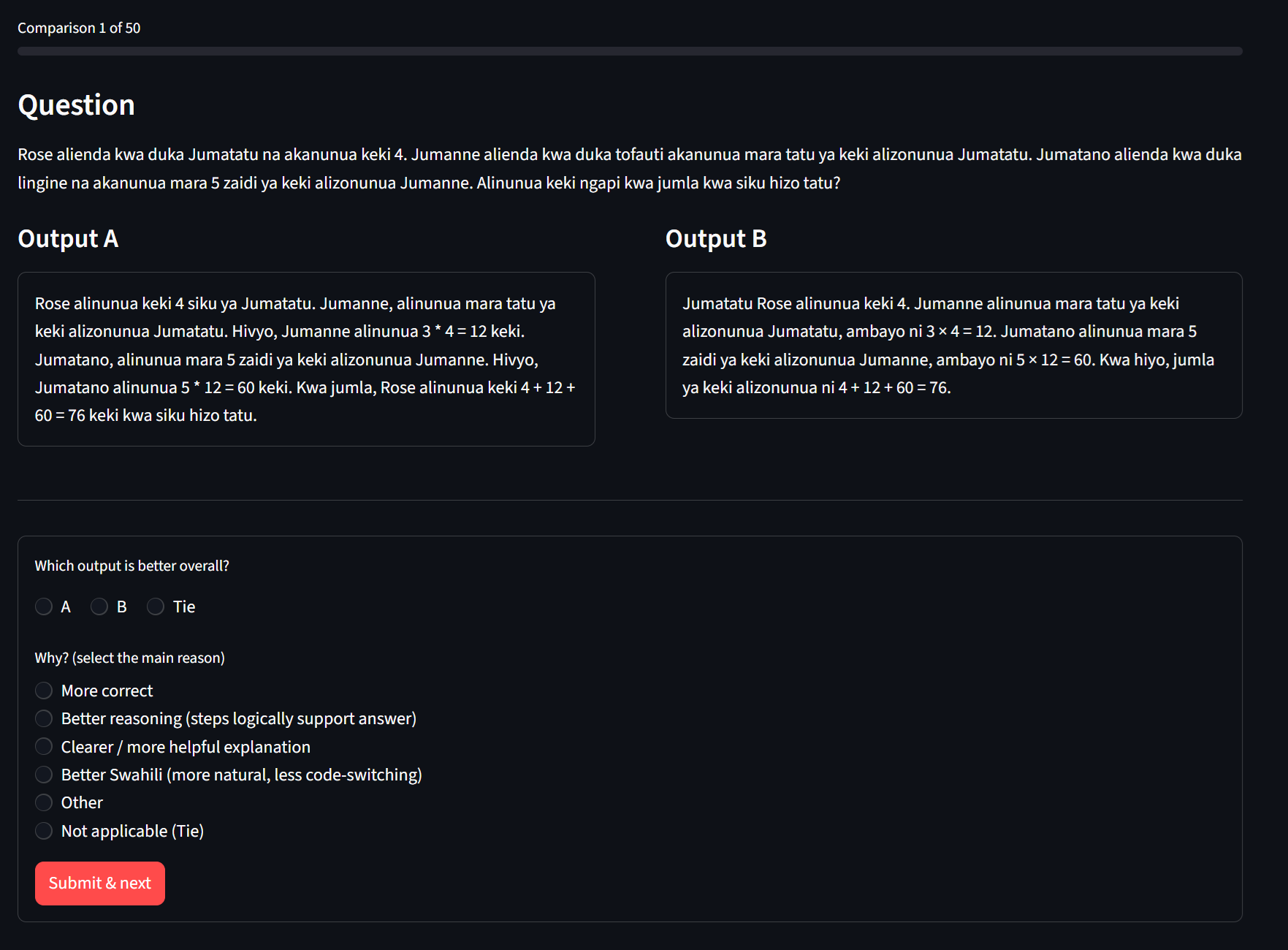}
    \caption{Screenshot of the tool used for our human evaluation (see Section~\ref{sec:native_speaker_protocol}).}
    \label{fig:human_eval_example}
\end{figure}

Figure~\ref{fig:human_eval_example} shows a screenshot of the tool used for human evaluation and described in Section \ref{sec:native_speaker_protocol}. Formatting (e.g., newlines) was removed from the text blocks and both annotators saw the same set of sample pairs drawn from the AfriMGSM test split. We used GPT-4o (gpt-4o-2024-08-06) for our comparison to Fikiria-14B. 

\section{Translation prompt templates}
\label{app:translation_templates}
\begin{promptbox}{Translation System Prompt}
You are a precise translation system.

Your task:
Translate the provided QUESTION and CHAIN-OF-THOUGHT (CoT) into the target language.

Requirements:
- Preserve meaning exactly.
- Do NOT summarize.
- Do NOT explain.
- Do NOT add commentary.
- Preserve logical structure of the reasoning.
- Output must strictly follow the specified format.
\end{promptbox}

\begin{promptbox}{Translation Prompt Template}
Translate the following content into {TARGET_LANGUAGE}.

You must:
- Translate faithfully.
- Preserve reasoning structure.
- Keep formatting identical.
- Output ONLY the translated result.
- Do not improve, refine, or correct the reasoning.
- Translate literally while preserving logical form.

FORMAT REQUIREMENTS:

<TRANSLATED_QUESTION>
...
</TRANSLATED_QUESTION>

<TRANSLATED_COT>
...
</TRANSLATED_COT>

{SHOTS}

NOW TRANSLATE THE FOLLOWING:

Input:

<QUESTION>
{QUESTION}
</QUESTION>

<COT>
{COT}
</COT>
\end{promptbox}

\begin{promptbox}{Shot Template}
EXAMPLE {N}

Input:

<QUESTION>
{QUESTION}
</QUESTION>

<COT>
{COT}
</COT>

Output:

<TRANSLATED_QUESTION>
{TRANSLATED_QUESTION}
</TRANSLATED_QUESTION>

<TRANSLATED_COT>
{TRANSLATED_COT}
</TRANSLATED_COT>

\end{promptbox}

\section{Evaluation prompt template}
\label{app:evaluation_templates}

\begin{promptbox}{Prompt template}
You are a helpful assistant that solves math word problems.
Think through the problem step by step in {language_name}, showing your calculations.

Put your reasoning inside <{rationale_tag}>...</{rationale_tag}> tags and your final numeric answer inside <answer>...</answer> tags.
\end{promptbox}

\end{document}